\documentclass{article}

\PassOptionsToPackage{numbers, compress}{natbib}

\usepackage[preprint]{neurips_2021}

\usepackage[utf8]{inputenc} 
\usepackage[T1]{fontenc}    
\usepackage{hyperref}       
\usepackage{url}            
\usepackage{booktabs}       
\usepackage{amsfonts}       
\usepackage{nicefrac}       
\usepackage{microtype}      

\usepackage{amssymb}
\usepackage{graphicx}
\usepackage{amsmath}
\usepackage[dvipsnames]{xcolor}
\usepackage{tikz}
\usetikzlibrary{calc}
\usetikzlibrary{positioning}
\usetikzlibrary{decorations.pathreplacing,angles,quotes}
\usepackage{float}
\usepackage{wrapfig}
\usepackage{subcaption}
\title{Convolutional Deep Exponential Families}

\author{%
  Chengkuan Hong, Christian R. Shelton\\
  Department of Computer Science and Engineering\\
  University of California, Riverside\\
  \texttt{chong009@ucr.edu, cshelton@cs.ucr.edu} \\
}

\begin{document}

\maketitle

\begin{abstract}
  We describe \textit{convolutional deep exponential families} (CDEFs) in this paper. CDEFs are built based on \textit{deep exponential families}, deep probabilistic models that capture the hierarchical dependence between latent variables. CDEFs greatly reduce the number of free parameters by tying the weights of DEFs. Our experiments show that CDEFs are able to uncover time correlations with a small amount of data.
\end{abstract}

\section{Introduction}

Deep neural networks (DNNs) \cite{schmidhuber2015deep} have achieved great success \cite{lecun2015deep} in many areas \cite{goodfellow2016deep}, including computer vision, natural language processing and time series analysis. Nevertheless, neural networks have limitations. As a supervised learning method, the testing data should be in the same form as the training data, while it is often not the case in practice, and does not allow for missing data. Additionally, a deep neural network needs a large amount of training data. 

Deep exponential families (DEFs) \cite{ranganath2015deep}, an unsupervised probabilistic graphical models \cite{koller2009probabilistic}, are a good supplement to deep learning. DEFs have the deep structure to learn the hierarchical information of the data. DEFs can also represent the data utilizing the power of exponential families. Unlike DNNs, they are able to predict any variable set based on any other, and are not limited to the input-output pattern in the training data. So a DEF is able to predict the missing information with any small part of the testing data.

In this paper, we develop convolutional deep exponential families (CDEFs), a modified model based on DEFs. A DEF, like a neural network, is composed of fully connected layers of latent variables. Both the variables in each layer and the variables in the connections are from exponential families.

In CDEFs, we tie the weights, like in a convolutional neural network \cite{lecun1995convolutional}, which dramatically reduces the number of parameters. Thereby, CDEFs are able to capture time correlations with less data.

\paragraph{Exponential families} An exponential family \cite{brown1986fundamentals} is a set of probability distributions that satisfy a specific form:$$p(x)=h(x)\exp(\eta^TT(x)-a(\eta)),$$
where $h$ is the base measure, $\eta$ are the natural parameters, $T$ are the sufficient statistics, and $a$ is the log-normalizer.

\paragraph{Deep exponential families} In order to construct deep exponential families, the latent layers of variables are stacked hierarchically. The parameters of each variable are controlled by the variables of the layer above and the connection weights. 

For a deep exponential family model, we have $L$ hidden layers $\{\mathbf{z}_{1}, \cdots, \mathbf{z}_{L}\}$ for each data point $x$. Each of the layers contains $K_\ell$ latent variables $\mathbf{z}_{\ell}=\{z_{\ell,1},\cdots, z_{\ell,K_\ell}\}$, where $z_{\ell,k}$ is assumed to be a scalar. This model contains $L-1$ layers of weights $\{\mathbf{W}_1,\cdots,\mathbf{W}_{L-1}\}$, which are shared across training and testing data. Each $\mathbf{W}_{\ell}$ is a $K_\ell \times K_{\ell+1}$ matrix. We assume there are prior distributions $p(\mathbf{W}_{\ell})$ for the weights.

\begin{wrapfigure}{l}{0.39\textwidth}
\resizebox{.35\textwidth}{!}{
\begin{tikzpicture}
[
recnode/.style={      rectangle,
      draw=black,
      thick,
      fill=white,
      text width=5cm,
      align=right,
      minimum height=1.25cm},
recnode2/.style={      rectangle,
      thick,
      text width=7.5cm,
      align=center,
      minimum height=1.25cm},   
recnode3/.style={      rectangle,
      draw=black,
      fill=white,
      thick,
      text width=7.5cm,
      minimum height=1.25cm},  
circlenode/.style={circle, draw=black, very thick, minimum size=6mm, inner sep=0pt},
circlenode2/.style={circle, very thick, minimum size=6mm, inner sep=0pt},
textnode1/.style={circle, very thick, minimum size=6mm, inner sep=0pt},
blacknode/.style={circle,draw=black, fill=black, very thick, minimum size=2mm, inner sep=0pt}
]
\node[recnode] (1) {};
\node[recnode2,below=0.625cm of 1.south east, anchor=north east] (2) {};
\node[recnode3,below=0.625cm of 2.south east, anchor=north east] (3) {};
\node[recnode3,below=0.625cm of 3.south east, anchor=north east] (4) {};
\node[recnode2,below=0.625cm of 4.south east, anchor=north east] (5) {};
\node[recnode3,below=0.625cm of 5.south east, anchor=north east] (6) {};
\node[recnode3,below=0.625cm of 6.south east, anchor=north east] (7) {};
\node[circlenode] at (3.center) (8) {};
\node[textnode1,right=0.05cm of 8.east] (15) {\LARGE \(\mathbf{z_{\ell+1,k}}\)};
\node[blacknode, left=1mm of 3.west] (22) {};
\node[textnode1,left=1.7cm of 8.west] (29) {\LARGE \(\mathbf{w}_{\ell+1,k}\)};

\node[circlenode2] at (2.center) (9) {\LARGE \(\vdots\)};

\node[circlenode,above=1.5625cm of 2.center] (10) {};
\node[textnode1,right=0.05cm of 10.east] (16) {\LARGE \(\mathbf{z}_{L,k}\)};
\node[textnode1,left=2.3cm of 10.west] (20) {\LARGE \(\eta \)};
\node[blacknode, right=0.01mm of 20.east] (21) {};

\node[circlenode] at (4.center) (11) {};
\node[textnode1,right=0.05cm of 11.east] (17) {\LARGE \(\mathbf{z}_{\ell,k}\)};
\node[blacknode, left=1mm of 4.west] (23) {};
\node[textnode1,left=1.7cm of 11.west] (26) {\LARGE \(\mathbf{w}_{\ell,k}\)};

\node[circlenode2] at (5.center) (12) {\LARGE \(\vdots\)};
\node[circlenode] at (6.center) (13) {};
\node[textnode1,right=0.05cm of 13.east] (18) {\LARGE \(\mathbf{z}_{1,k}\)};
\node[blacknode, left=1mm of 6.west] (24) {};
\node[textnode1,left=2cm of 13.west] (28) {\LARGE \(\mathbf{w}_{1,k}\)};

\node[circlenode,fill=black!10] at (7.center) (14) {};
\node[textnode1,right=0.05cm of 14.east] (19) {\LARGE \(\mathbf{x}_{i}\)};
\node[blacknode, left=1mm of 7.west] (25) {};
\node[textnode1,left=2cm of 14.west] (27) {\LARGE \(\mathbf{w}_{0,i}\)};

\node[circlenode,left=1cm of 8.west] (30) {};
\node[circlenode,left=1cm of 11.west] (31) {};
\node[circlenode,left=1cm of 13.west] (32) {};
\node[circlenode,left=1cm of 14.west] (33) {};
\draw[->,line width=0.4mm] (30.east) -- (8.west);
\draw[->,line width=0.4mm] (31.east) -- (11.west);
\draw[->,line width=0.4mm] (32.east) -- (13.west);
\draw[->,line width=0.4mm] (33.east) -- (14.west);

\draw[->,line width=0.4mm] (21.east) -- (10.west);
\draw[->,line width=0.4mm] (10.south) -- (9.north);
\draw[->,line width=0.4mm] (9.south) -- (8.north);
\draw[->,line width=0.4mm] (8.south) -- (11.north);
\draw[->,line width=0.4mm] (11.south) -- (12.north);
\draw[->,line width=0.4mm] (12.south) -- (13.north);
\draw[->,line width=0.4mm] (13.south) -- (14.north);

\draw[->,line width=0.4mm,bend right=90] (22.south east) to [out=-30,in=-150] (30.south west);
\draw[->,line width=0.4mm,bend right=90] (23.south east) to [out=-30,in=-150] (31.south west);
\draw[->,line width=0.4mm,bend right=90] (24.south east) to [out=-30,in=-150] (32.south west);
\draw[->,line width=0.4mm,bend right=90] (25.south east) to [out=-30,in=-150] (33.south west);

\node[textnode1, above left=0.2mm and 1 mm of 1.south east] {\LARGE \(K_L\)};
\node[textnode1, above left=0.2mm and 1.7 mm of 3.south east] {\LARGE \(K_{\ell+1}\)};
\node[textnode1, above left=0.2mm and 1 mm of 4.south east] {\LARGE \(K_{\ell}\)};
\node[textnode1, above left=0.2mm and 1 mm of 6.south east] {\LARGE \(K_{1}\)};
\node[textnode1, above left=0.2mm and 1 mm of 7.south east] {\LARGE \(V\)};
\end{tikzpicture}
}
\caption{The structure of deep exponential families with $V$ observations for data point \(x\). \(x_i\) represents the \(i\)th observation. Reproduced based on Figure 2 of \cite{ranganath2015deep}.} 
\label{fig:defstructure}
\vskip -0.2in
\end{wrapfigure}
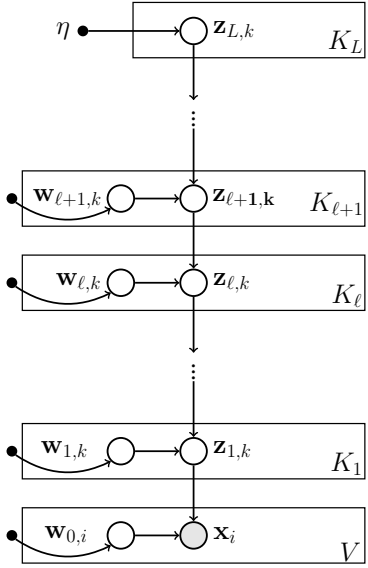

As shown in Figure \ref{fig:defstructure}, the top layer of latent variables are drawn from an exponential family:$$p(z_{L,k})=\textrm{EXPFAM}_L(z_{L,k},\eta),$$ where $\eta$ is a given parameter, and $\textrm{EXPFAM}(x,\eta)$ denotes $x$ is drawn from an exponential family with natural parameter $\eta$.

Next, each latent variable is drawn conditional on the previous layer: $$p(z_{\ell,k}|\mathbf{z}_{\ell+1}, \mathbf{w}_{\ell,k})=\textrm{EXPFAM}_\ell(z_{\ell,k},\text{g}_\ell(\mathbf{z}_{\ell+1}^T\mathbf{w}_{\ell,k})),$$ where $\text{g}_\ell$, called the \textit{link function}, maps the inner product to the natural parameter, $z_{\ell,k}$ is a scalar, $\mathbf{z}_{\ell+1}$ is a $K_{\ell+1}$ vector and $\mathbf{w}_{\ell,k}$ is a row vector from the $K_{\ell} \times K_{\ell+1}$ matrix $\mathbf{W}_\ell$. 

\paragraph{Convolutional Deep Exponential Families} Like the convolutional neural network, the weights of the convolutional deep exponential families are also tied such that the weight matrix only has small number of free parameters.

For example, for a $5 \times 3$ matrix $\mathbf{W}$, in the original deep exponential families model, $\mathbf{W}_{original}$ has 15 free parameters. But, for a convolutional deep exponential families model, $\mathbf{W}_{conv}$ only has 3 free parameters when the filter size is 3 and the stride is 1: 
\vskip .2in

\begin{equation}
\mathbf{W}_{original}=
\begin{bmatrix}
w_{11} & w_{12} & w_{13}\\
\vdots & \vdots & \vdots \\
w_{51} & w_{52} & w_{53}
\end{bmatrix},\ 
\mathbf{W}_{conv}=\begin{bmatrix}
w_{11} & 0 & 0\\
w_{21} & w_{11} & 0\\
w_{31} & w_{21} & w_{11}\\
0 & w_{31} & w_{21} \\
0 & 0 & w_{31}
\end{bmatrix}
\end{equation}

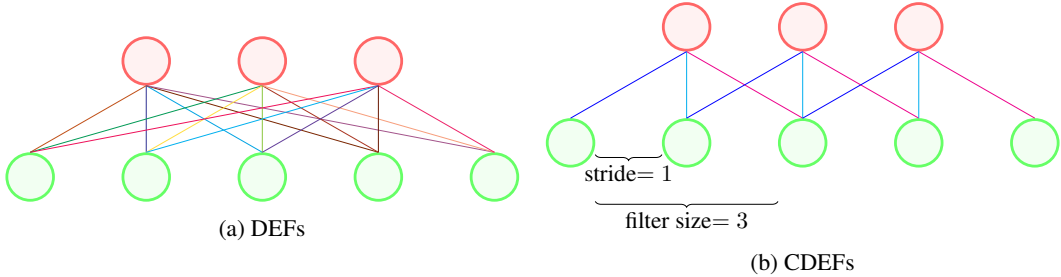
\begin{figure}[H]
\begin{subfigure}[h]{.5\textwidth}
\resizebox{\columnwidth}{!}{
\begin{tikzpicture}[
roundnode/.style={circle, draw=green!60, fill=green!5, very thick, minimum size=7mm},
squarednode/.style={circle, draw=red!60, fill=red!5, very thick, minimum size=7mm}
]
\node[squarednode]      (upper1)                              {};
\node[squarednode]      (upper2)        [right=of upper1]        {};
\node[squarednode]      (upper3)        [right=of upper2]        {};
\node[roundnode]        (lower2)       [below=of upper1] {};
\node[roundnode]        (lower1)        [left=of lower2]       {};
\node[roundnode]        (lower3)       [below=of upper2] {};
\node[roundnode]        (lower4)       [below=of upper3] {};
\node[roundnode]        (lower5)       [right=of lower4]  {};
 
\draw[-, Bittersweet] (upper1.south) -- (lower1.north);
\draw[-, Blue] (upper1.south) -- (lower2.north);
\draw[-, Cerulean] (upper1.south) -- (lower3.north);
\draw[-, Brown] (upper1.south) -- (lower4.north);
\draw[-,DarkOrchid] (upper1.south) -- (lower5.north);

\draw[-,ForestGreen] (upper2.south) -- (lower1.north);
\draw[-, Goldenrod] (upper2.south) -- (lower2.north);
\draw[-, LimeGreen] (upper2.south) -- (lower3.north);
\draw[-, Mahogany] (upper2.south) -- (lower4.north);
\draw[-, Melon] (upper2.south) -- (lower5.north);

\draw[-,OrangeRed] (upper3.south) -- (lower1.north);
\draw[-,ProcessBlue] (upper3.south) -- (lower2.north);
\draw[-,RoyalPurple] (upper3.south) -- (lower3.north);
\draw[-,Sepia] (upper3.south) -- (lower4.north);
\draw[-,WildStrawberry] (upper3.south) -- (lower5.north);

\end{tikzpicture}
}
\caption{DEFs}
\label{fig:defs}
\end{subfigure}
~
\begin{subfigure}[h]{.5\textwidth}
\resizebox{\columnwidth}{!}{
\begin{tikzpicture}[
roundnode/.style={circle, draw=green!60, fill=green!5, very thick, minimum size=7mm},
squarednode/.style={circle, draw=red!60, fill=red!5, very thick, minimum size=7mm},
]
\node[squarednode]      (upper1)                              {};
\node[squarednode]      (upper2)        [right=of upper1]        {};
\node[squarednode]      (upper3)        [right=of upper2]        {};
\node[roundnode]        (lower2)       [below=of upper1] {};
\node[roundnode]        (lower1)        [left=of lower2]       {};
\node[roundnode]        (lower3)       [below=of upper2] {};
\node[roundnode]        (lower4)       [below=of upper3] {};
\node[roundnode]        (lower5)       [right=of lower4]  {};
 
\draw[-,blue] (upper1.south) -- (lower1.north);
\draw[-,cyan] (upper1.south) -- (lower2.north);
\draw[-, magenta] (upper1.south) -- (lower3.north);

\draw[-,blue] (upper2.south) -- (lower2.north);
\draw[-,cyan] (upper2.south) -- (lower3.north);
\draw[-, magenta] (upper2.south) -- (lower4.north);

\draw[-,blue] (upper3.south) -- (lower3.north);
\draw[-,cyan] (upper3.south) -- (lower4.north);
\draw[-, magenta] (upper3.south) -- (lower5.north);
\draw[decoration={brace,mirror,raise=5pt},decorate]
  (lower1) -- node[below=6pt] {stride$=1$} (lower2);
  \draw[decoration={brace,mirror,raise=25pt},decorate]
  (lower1) -- node[below=26pt] {filter size$=3$} (lower3);
\end{tikzpicture}
}
\caption{CDEFs}
\label{fig:cdefs}
\end{subfigure}
\caption{DEFs vs. CDEFs}
\label{fig:defvscdef}
\end{figure}

Figure \ref{fig:defvscdef} shows the connections for DEFs and CDEFs. Figure \ref{fig:defs} shows 15 different weights while Figure \ref{fig:cdefs} only has 3 different weights, where the same color represents the same weights. 

\paragraph{Likelihood} The observation data is assumed to be drawn conditioned on the lowest hidden layer of the DEF, $p(x_{n,i}|\mathbf{z}_{n,1})$. Since we focus on count data, we use the Poisson distribution as the distribution for the observation data.

If we let $x_{n,i}$ be the count of event $i$ in sample $n$ and $\mathbf{z_{n,1}}$ is the corresponding hidden variable in the first layer, then the likelihood of $x_{n,i}$ would be $$p(x=x_{n,i}|\mathbf{z_{n,1}}, \mathbf{W_0}) = \text{Poisson}(x_{n,i},\lambda=\mathbf{z_{n,1}}^T\mathbf{w_{0,i}})=e^{-\lambda}\frac{\lambda^x}{x!}.$$ The elements of the observation matrix $\mathbf{W}_0$ are from gamma distributions and also tied to be a convolutional matrix.

\section{Convolutional Sparse Gamma DEF}
In this paper, we implemented a convolutional sparse gamma DEF.

The sparse gamma DEF is a DEF with gamma-distributed latent layers. The probability density of the gamma distribution is $$p(z)=z^{-1}\exp(\alpha \log(z) - \beta z - \log \Gamma (\alpha) + \alpha \log(\beta)),$$ where $\alpha$ and $\beta$ are natural parameters and $\Gamma$ is the gamma function.

The parameters of a layer are controlled by its immediately higher layer and the weights through the link function, which maps the inner product of the hidden layer and the weights to the parameter of the layer. The link function is given as $$\text{g}_\alpha = \alpha_\ell,\ \text{g}_\beta=\frac{\alpha_\ell}{\mathbf{z^T_{\ell+1}w_{\ell,k}}}$$ 
From the link function, we can see the shape is fixed for all the layers while the scale is modified to control the expectation, $E(z)=\alpha \beta^{-1}$.

The shape of the weights and hidden layers are set to be less than 1. This kind of gamma distribution is called a \textit{soft gamma}. Most data points drawn from this type of distribution are near 0. It has shown great performance on feature selection and unsupervised feature discovery \cite{Goodfellow+al-ICML2012,hernandez2013generalized}. 
\section{Inference}
To update the parameters of a CDEF, we need to solve the posterior inference problem. Here, we used black box variational inference \cite{ranganath2014black} for the posterior inference.

Variational inference \cite{jordan1999introduction} seeks to solve an optimization problem. It aims to minimize the KL divergence from an approximating distribution to the posterior, which is equivalent to maximizing the Evidence Lower Bound(ELBO)\cite{bishop:2006:PRML}: $$\mathcal{L}(q)=E_{q(\mathbf{z,W})}[\log p(\mathbf{x,z,W}) - \log q(\mathbf{z,W})],$$ where $\mathbf{z}$ denotes all the latent variables and $\mathbf{W}$ denotes the weights. This function is the lower bound on $\log p(x)$, which we will maximize by gradient ascent.

The approximating distribution $q$ is assumed to be in the mean field variational family. Under the mean field assumption, $$q(\mathbf{z,W})=q(\mathbf{W_0})\prod^L_{\ell=1}q(\mathbf{W_\ell})\prod^N_{n=1}q(\mathbf{z_{n,\ell}}),$$ where $q(\mathbf{z_{n,\ell}})$ and $q(\mathbf{W_\ell)}$ are fully factorized, $n$ is the sample index and $\ell$ is the layer index. We have a different hidden variable $\mathbf{z_{n,\ell}}$ for a different sample $x_n$.

Each component in $q(\mathbf{z_{n,\ell}})$ is $$q(z_{n,\ell,k})=\text{EXPFAM}_\ell(z_{n,\ell,k}, \lambda_{n,\ell,k}),$$ where $q(\mathbf{z})$ and $p(\mathbf{z})$ are in the same exponential family, $z_{n,\ell,k}$ is the $k^{th}$ hidden variable in layer $\ell$ for sample $n$, and $\lambda_{n,\ell,k}$ is the corresponding parameter.

$q(\mathbf{W})$ and $p(\mathbf{W})$ are also from the same exponential family, with parameter $\xi$.

Let $p_{n,\ell,k}(\mathbf{x,z,W})$ be the probability of the Markov blanket that contains $z_{n,\ell,k}$. Then, the gradient for the approximation of $z_{n,\ell,k}$ is $$\nabla_{\lambda_{n,\ell,k}}\mathcal{L}=E_{\mathbf{z}\sim q(\mathbf{z})}[[\nabla_{\lambda_{n,\ell,k}}\log q(z_{n,\ell,k})](\log p_{n,\ell,k}(\mathbf{x,z,W})-\log q(z_{n,\ell,k})].$$ 
For the original DEFs, the probability of the Markov blanket for a latent variable in the first layer is $$\log p_{n,1,k}(\mathbf{x},\mathbf{z},\mathbf{W})=\log p(z_{n,1,k}|\mathbf{z_{n,2}}, \mathbf{w_{1,k}}) + \log p(\mathbf{x_{n}}| \mathbf{z_{n,1}}, \mathbf{W_0}).$$
For CDEFs, it becomes $$\log p_{n,1,k}(\mathbf{x},\mathbf{z},\mathbf{W})=\log p(z_{n,1,k}|\mathbf{z_{n,2}}, \mathbf{w_{1,k}}) + \log p(\mathbf{x_{n_k}}| z_{n,1,k}, \mathbf{W_0}),$$
where $\mathbf{x_{n_k}}$ denotes the observations connected to the hidden node $z_{n,1,k}$.

In DEFs, the probability of the Markov blanket for a latent variable in the intermediate layer is $$\log p_{n,\ell,k}(\mathbf{x},\mathbf{z},\mathbf{W})=\log p(z_{n,\ell,k}|\mathbf{z_{n,\ell+1}}, \mathbf{w_{\ell,k}}) + \log p(\mathbf{z_{n,\ell-1}}| \mathbf{z_{n,\ell}}, \mathbf{W_{\ell-1}}),$$
while for CDEFs, it becomes $$\log p_{n,\ell,k}(\mathbf{x},\mathbf{z},\mathbf{W})=\log p(z_{n,\ell,k}|\mathbf{z_{n,\ell+1}}, \mathbf{w_{\ell,k}}) + \log p(\mathbf{z_{n_k,\ell-1}}| z_{n,\ell,k}, \mathbf{W_{\ell-1}}),$$
where $\mathbf{z_{n_k,\ell-1}}$ denotes the hidden variables in the layer $\ell-1$ connected to the hidden node $z_{n,\ell,k}$.

In DEFs, the probability of the Markov blanket for the hidden variable in the top layer is $$\log p_{n,L,k}(\mathbf{x},\mathbf{z},\mathbf{W})=\log p(z_{n,L,k}) + \log p(\mathbf{z_{n,L-1}}| \mathbf{z_{n,L}}, \mathbf{W_{L-1}}).$$
For CDEFs, it becomes $$\log p_{n,L,k}(\mathbf{x},\mathbf{z},\mathbf{W})=\log p(z_{n,L,k}) + \log p(\mathbf{z_{n_k,L-1}}| z_{n,L,k}, \mathbf{W_{L-1}}),$$
where $\mathbf{z_{n_k,L-1}}$ denotes the hidden variables in the layer $L-1$ connected to the hidden node $z_{n,L,k}$ and $\log p(z_{n,L,k})$ is from a given prior distribution.

Not only did we optimize the ELBO with respect to the hidden variables $\mathbf{z}$, we also optimized the ELBO with respect to the weights $\mathbf{W}$ in the training process. The weights $\mathbf{W}$ were fixed in the testing process. We only updated the hidden variables $\mathbf{z}$ in the testing process.

Similarly, the gradient for $\mathbf{W}$ is $$\nabla_{\xi_{\ell,i,j}}\mathcal{L}=E_{\mathbf{W} \sim q(\mathbf{W}) }[[\nabla_{\xi_{\ell,i,j}}\log q(W_{\ell,i,j})](\log p_{\ell,i,j}(\mathbf{x},\mathbf{z},\mathbf{W})-\log q(W_{\ell,i,j}))],$$ where $W_{\ell,i,j}$ denotes the $(i,j)^{th}$ element of $\mathbf{W_\ell}$ and $p_{\ell,i,j}(\mathbf{x,z,W})$ is the probability of the Markov blanket that contains $W_{\ell,i,j}$. The only difference between DEFs and CDEFs is the term $\log p_{\ell,i,j}(\mathbf{x},\mathbf{z},\mathbf{W})$.

In DEFs, every $W_{\ell,i,j}$ corresponds to different $\xi_{\ell,i,j}$. While in CDEFs, several entries of $\mathbf{W_\ell}$ share a same parameter $\xi$.

In DEFs, the probability of the Markov blanket for $\mathbf{W_0}$ is $$\log p_{0,i,j}(\mathbf{x},\mathbf{z},\mathbf{W})=\log p(\mathbf{W_0})_{(i,j)} + \sum_{n=1}^N\log p(x_{n,i}| \mathbf{z_{n,1}}, \mathbf{W_0}).$$

In CDEFs, suppose $\mathbf{W_0}_{(i_1,j_1)}, \mathbf{W_0}_{(i_2,j_2)}, \cdots, \mathbf{W_0}_{(i_t,j_t)}$ share the same parameter $\xi_{0,i,j}$, then  the probability of the Markov blanket becomes $$\log p_{0,i,j}(\mathbf{x},\mathbf{z},\mathbf{W})=\log p(\mathbf{W_0})_{(i_1,j_1)} + \sum_{k=1}^t\sum_{n=1}^N\log p(x_{n,i_k}| \mathbf{z_{n,1}}, \mathbf{W_0}).$$

In DEFs, the probability of the Markov blanket for $\mathbf{W}_\ell$, where $\ell=\{1,\cdots,L-1\}$ is $$\log p_{\ell,i,j}(\mathbf{x},\mathbf{z},\mathbf{W})=\log p(\mathbf{W_\ell})_{(i,j)} + \sum_{n=1}^N\log p(z_{n,\ell,i}| \mathbf{z_{n,\ell+1}}, \mathbf{W_{\ell}}).$$

In CDEFs, suppose $\mathbf{W_\ell}_{(i_1,j_1)}, \mathbf{W_\ell}_{(i_2,j_2)}, \cdots, \mathbf{W_\ell}_{(i_t,j_t)}$ share the same parameter $\xi_{\ell,i,j}$, then the probability of the Markov blanket becomes $$\log p_{\ell,i,j}(\mathbf{x},\mathbf{z},\mathbf{W})=\log p(\mathbf{W_\ell})_{(i_1,j_1)} + \sum_{k=1}^t\sum_{n=1}^N\log p(z_{n,\ell,i_k}| \mathbf{z_{n,\ell+1}}, \mathbf{W_{\ell}}).$$

\section{Experiments}
We collected the crime data for Chicago from 2003 to 2016\footnote{https://data.cityofchicago.org/Public-Safety/Crimes-2001-to-present/ijzp-q8t2/data}.
The days of a year are truncated to 357, \textit{i.e.} 51 weeks. The days in a week start on Sunday.

\begin{figure}[h]
\resizebox{\columnwidth}{!}{
\begin{tikzpicture}[
roundnode/.style={circle, draw=green!60, fill=green!5, very thick, minimum size=1mm},
squarednode/.style={circle, draw=red!60, fill=red!5, very thick, minimum size=1mm},
]
\node[squarednode]      (1)                              {};
\node[squarednode]      (2)        [right=of 1]        {};
\node[]                 (3)        [right=of 2]        {$\cdots$};
\node[squarednode]      (4)        [right=of 3]        {};
\node[]                 (5)        [right=of 4]        {$\cdots$};
\node[squarednode]      (6)        [right=of 5]        {};
\node[squarednode]      (7)        [right=of 6]        {};
\node[]                 (8)        [right=of 7]        {$\cdots$};
\node[squarednode]      (9)        [right=of 8]        {};

\draw[decoration={brace,mirror,raise=6pt},decorate]
  (1) -- node[below=7pt] {77 locations for 1 day} (4);
\draw[decoration={brace,mirror,raise=6pt},decorate]
(6) -- node[below=7pt] {77 locations for 1 day} (9);
  \draw[decoration={brace,mirror,raise=20pt},decorate]
  (1) -- node[below=21pt] {357 days} (9);
\end{tikzpicture}
}
\caption{The representation of the data.}
\label{fig:datalocations}
\end{figure}
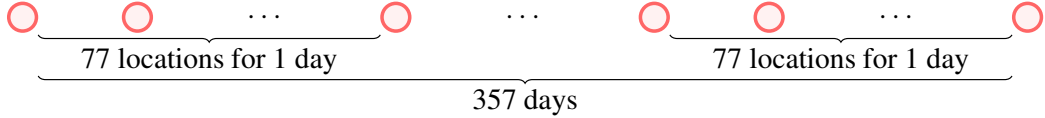

The data is arranged in the order: the number of thefts for each location for each day. There are 77 locations in Chicago. So we have 14 samples (14 years), each of which has 27489 numbers, representing the numbers of thefts for each location for each day in that year. As in Figure \ref{fig:datalocations}, each node represents the number of crimes for that location.

The models we used are all convolutional sparse gamma DEF. 
\subsection{The Benefit of The Second Layer}
We constructed 5 models, drawn in Figure \ref{fig:cdefsstructurelocation}, to compare the results of test log likelihood:
\begin{itemize}
\item Homogeneous Poisson process (HP): Use the maximum likelihood estimation to estimate the rates of thefts for each location.
\item CDEFs 1-51: 1 layer CDEF with 51 hidden variables in the hidden layer. There is no overlap between filters. Each hidden node is connected with all the observed data in 1 week. The filter size is 539 ( $7 \text{ days }\times 77\text{ locations}$) and the stride is also 539. 
\item CDEFs 1-51, 2-17: 2 layers CDEF with 51 hidden variables in the first hidden layer and 17 hidden nodes in the second hidden layer. The first hidden layer is the same as CDEFs 1-51. Each node of the second layer is connected with 3 hidden nodes in the first hidden layer. The filter size for the second layer is 3 and the stride is 3.
\item CDEFs 1-51, 2-25: 2 layers CDEFs with 51 hidden nodes in the first hidden layer, the same as CDEFs 1-51, and 25 hidden nodes in the second hidden layer. The filter size for the second hidden layer is 3 and stride is 2.
\item CDEFs 1-51, 2-49: 2 layers CDEFs  with 51 hidden nodes in the first hidden layer, the same as CDEFs 1-51, and 49 hidden nodes in the second hidden layer. The filter size for the second hidden layer is 3 and stride is 1.
\end{itemize}
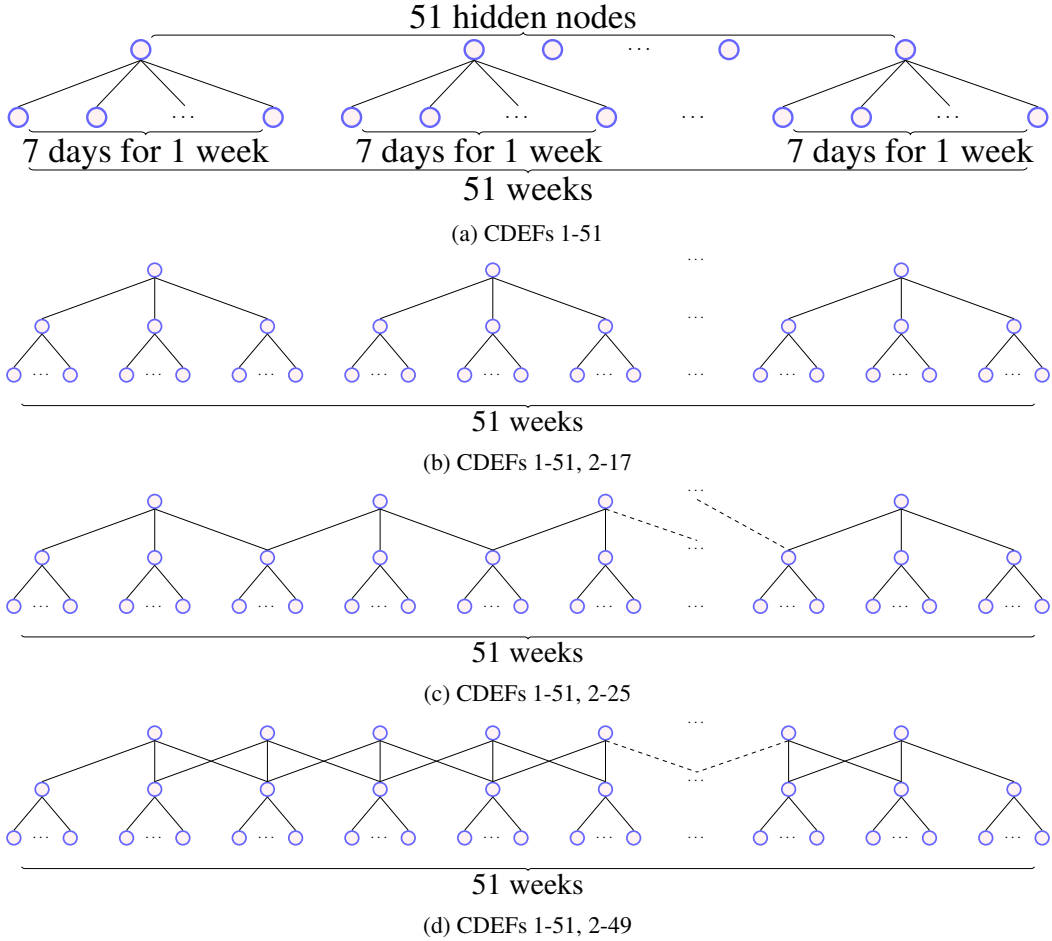
\begin{figure}[H]
\centering
\begin{subfigure}[h]{\textwidth}
\resizebox{\columnwidth}{!}{
\begin{tikzpicture}[
roundnode/.style={circle, draw=green!60, fill=green!5, very thick, minimum size=1mm},
squarednode/.style={circle, draw=blue!60, fill=red!5, very thick, minimum size=1mm},
]
\node[squarednode]      (1)                              {};
\node[squarednode]      (2)        [right=of 1]        {};
\node[]                 (3)        [right=of 2]        {$\cdots$};
\coordinate (Middle) at ($(2)!0.5!(3)$);
\node[squarednode]      (10)       [above=of Middle]   {};
\node[squarednode]      (4)        [right=of 3]        {};

\node[squarednode]      (16)        [right=of 4]                      {};
\node[squarednode]      (17)        [right=of 16]        {};
\node[]                 (18)        [right=of 17]        {$\cdots$};
\coordinate (Middle3) at ($(17)!0.5!(18)$);
\node[squarednode]      (19)       [right=of 18]   {};

\node[]                 (5)        [right=of 19]        {$\cdots$};
\node[squarednode]      (6)        [right=of 5]        {};
\node[squarednode]      (7)        [right=of 6]        {};
\node[]                 (8)        [right=of 7]        {$\cdots$};
\coordinate (Middle2) at ($(7)!0.5!(8)$);
\node[squarednode]      (11)       [above=of Middle2]   {};
\node[squarednode]      (9)        [right=of 8]        {};

\node[squarednode]      (12)        [above=of Middle3]        {};
\node[squarednode]      (13)        [right=of 12]        {};
\node[]      (14)        [right=of 13]        {$\cdots$};
\node[squarednode]      (15)        [right=of 14]        {};

\draw[-,black] (10.south) -- (1.north);
\draw[-,black] (10.south) -- (2.north);
\draw[-,black] (10.south) -- (3.north);
\draw[-, black] (10.south) -- (4.north);

\draw[-,black] (11.south) -- (6.north);
\draw[-,black] (11.south) -- (7.north);
\draw[-,black] (11.south) -- (8.north);
\draw[-, black] (11.south) -- (9.north);

\draw[-,black] (12.south) -- (16.north);
\draw[-,black] (12.south) -- (17.north);
\draw[-,black] (12.south) -- (18.north);
\draw[-, black] (12.south) -- (19.north);

\draw[decoration={brace,mirror,raise=6pt},decorate]
  (1) -- node[below=7pt,font=\LARGE] {7 days for 1 week} (4);
\draw[decoration={brace,mirror,raise=6pt},decorate]
(6) -- node[below=7pt,font=\LARGE] {7 days for 1 week} (9);
\draw[decoration={brace,mirror,raise=6pt},decorate]
(16) -- node[below=7pt,font=\LARGE] {7 days for 1 week} (19);
\draw[decoration={brace,raise=6pt},decorate]
(10) -- node[above=7pt,font=\LARGE] {51 hidden nodes} (11);
  \draw[decoration={brace,mirror,raise=25pt},decorate]
  (1) -- node[below=26pt,font=\LARGE] {51 weeks} (9);
\end{tikzpicture}
}
\caption{CDEFs 1-51}
\end{subfigure}
\hfill
\begin{subfigure}[h]{\textwidth}
\resizebox{\columnwidth}{!}{
\begin{tikzpicture}[
roundnode/.style={circle, draw=green!60, fill=green!5, very thick, minimum size=1mm},
squarednode/.style={circle, draw=blue!60, fill=red!5, very thick, minimum size=1mm},
]
\node[squarednode]      (1)                              {};
\node[squarednode]      (3)        [right=of 1]        {};
\coordinate (2) at ($(1)!0.5!(3)$);
\draw (2) node[] {$\cdots$};

\node[squarednode]      (4)        [right=of 3]                      {};
\node[squarednode]      (6)       [right=of 4]   {};
\coordinate (5) at ($(4)!0.5!(6)$);
\draw (5) node[] {$\cdots$};

\node[squarednode]      (7)        [right=of 6]        {};
\node[squarednode]      (9)        [right=of 7]        {};
\coordinate (8) at ($(7)!0.5!(9)$);
\draw (8) node[] {$\cdots$};

\node[squarednode]      (10)	[right=of 9]           {};
\node[squarednode]      (12)        [right=of 10]        {};
\coordinate (11) at ($(10)!0.5!(12)$);
\draw (11) node[] {$\cdots$};

\node[squarednode]      (13)        [right=of 12]                      {};
\node[squarednode]      (15)       [right=of 13]   {};
\coordinate (14) at ($(13)!0.5!(15)$);
\draw (14) node[] {$\cdots$};

\node[squarednode]      (16)        [right=of 15]        {};
\node[squarednode]      (18)        [right=of 16]        {};
\coordinate (17) at ($(16)!0.5!(18)$);
\draw (17) node[] {$\cdots$};

\node[]      (19)        [right=of 18]        {$\cdots$};

\node[squarednode]      (20)       [right=of 19]   {};
\node[squarednode]      (22)        [right=of 20]        {};
\coordinate (21) at ($(20)!0.5!(22)$);
\draw (21) node[] {$\cdots$};

\node[squarednode]      (23)       [right=of 22]   {};
\node[squarednode]      (25)        [right=of 23]        {};
\coordinate (24) at ($(23)!0.5!(25)$);
\draw (24) node[] {$\cdots$};

\node[squarednode]      (26)       [right=of 25]   {};
\node[squarednode]      (28)        [right=of 26]        {};
\coordinate (27) at ($(26)!0.5!(28)$);
\draw (27) node[] {$\cdots$};

\foreach \x in {29,...,38}{ 
	\pgfmathsetmacro\indexo{\x-28}
    \pgfmathsetmacro\indext{3*\indexo-1}
    \pgfmathsetmacro\indextt{\indext-2}
    \ifnum\x=35
    	\node[] (\x) [above=of 19] {$\cdots$};
    \else
    	\ifnum\x<35
			\node[squarednode] (\x) [above=of \indext] {};
        \else
        	\node[squarednode] (\x) [above=of \indextt] {};
        \fi
    \fi    
}   
\foreach \x in {29,...,38}{ 
	\pgfmathsetmacro\indexo{\x-28}
    \pgfmathsetmacro\indext{3*\indexo-1}
    \pgfmathsetmacro\indextt{\indext-2}
            	\pgfmathtruncatemacro\lc{\indext-1}
            \pgfmathtruncatemacro\rc{\indext+1}
            \pgfmathtruncatemacro\lct{\indextt-1}
            \pgfmathtruncatemacro\rct{\indextt+1}
    \ifnum\x=35

    \else
    	\ifnum\x<35
        	\draw[-,black] (\x.south) -- (\lc.north);
            \draw[-,black] (\x.south) -- (\rc.north);
        \else
        	\draw[-,black] (\x.south) -- (\lct.north);
            \draw[-,black] (\x.south) -- (\rct.north);
        \fi
    \fi    
} 

\foreach \x in {39,...,42}{ 
	\pgfmathsetmacro\indexo{\x-38}
    \pgfmathtruncatemacro\indext{3*\indexo+27}
    \pgfmathtruncatemacro\indextt{\indext-2}
            	\pgfmathtruncatemacro\lc{\indext-1}
            \pgfmathtruncatemacro\rc{\indext+1}
            \pgfmathtruncatemacro\lct{\indextt-1}
            \pgfmathtruncatemacro\rct{\indextt+1}
    \ifnum\x<41
    		\node[squarednode] (\x) [above=of \indext] {};
        	\draw[-,black] (\x.south) -- (\lc.north);
            \draw[-,black] (\x.south) -- (\rc.north);
            \draw[-,black] (\x.south) -- (\indext.north);
    \else
    	\ifnum\x=41
        	\node[] (\x) [above=of 35] {$\cdots$};
        \else
        	\node[squarednode] (\x) [above=of \indextt] {};
            \draw[-,black] (\x.south) -- (\lct.north);
            \draw[-,black] (\x.south) -- (\rct.north);
            \draw[-,black] (\x.south) -- (\indextt.north);
        \fi
    \fi    
} 
\draw[decoration={brace,mirror,raise=20pt},decorate] (1) -- node[below=21pt, font=\huge] {51 weeks} (28);
\end{tikzpicture}
}
\caption{CDEFs 1-51, 2-17}
\end{subfigure}
\hfill

\begin{subfigure}[h]{\textwidth}
\resizebox{\columnwidth}{!}{
\begin{tikzpicture}[
roundnode/.style={circle, draw=green!60, fill=green!5, very thick, minimum size=1mm},
squarednode/.style={circle, draw=blue!60, fill=red!5, very thick, minimum size=1mm},
]
\node[squarednode]      (1)                              {};
\node[squarednode]      (3)        [right=of 1]        {};
\coordinate (2) at ($(1)!0.5!(3)$);
\draw (2) node[] {$\cdots$};

\node[squarednode]      (4)        [right=of 3]                      {};
\node[squarednode]      (6)       [right=of 4]   {};
\coordinate (5) at ($(4)!0.5!(6)$);
\draw (5) node[] {$\cdots$};

\node[squarednode]      (7)        [right=of 6]        {};
\node[squarednode]      (9)        [right=of 7]        {};
\coordinate (8) at ($(7)!0.5!(9)$);
\draw (8) node[] {$\cdots$};

\node[squarednode]      (10)	[right=of 9]           {};
\node[squarednode]      (12)        [right=of 10]        {};
\coordinate (11) at ($(10)!0.5!(12)$);
\draw (11) node[] {$\cdots$};

\node[squarednode]      (13)        [right=of 12]                      {};
\node[squarednode]      (15)       [right=of 13]   {};
\coordinate (14) at ($(13)!0.5!(15)$);
\draw (14) node[] {$\cdots$};

\node[squarednode]      (16)        [right=of 15]        {};
\node[squarednode]      (18)        [right=of 16]        {};
\coordinate (17) at ($(16)!0.5!(18)$);
\draw (17) node[] {$\cdots$};

\node[]      (19)        [right=of 18]        {$\cdots$};

\node[squarednode]      (20)       [right=of 19]   {};
\node[squarednode]      (22)        [right=of 20]        {};
\coordinate (21) at ($(20)!0.5!(22)$);
\draw (21) node[] {$\cdots$};

\node[squarednode]      (23)       [right=of 22]   {};
\node[squarednode]      (25)        [right=of 23]        {};
\coordinate (24) at ($(23)!0.5!(25)$);
\draw (24) node[] {$\cdots$};

\node[squarednode]      (26)       [right=of 25]   {};
\node[squarednode]      (28)        [right=of 26]        {};
\coordinate (27) at ($(26)!0.5!(28)$);
\draw (27) node[] {$\cdots$};

\foreach \x in {29,...,38}{ 
	\pgfmathsetmacro\indexo{\x-28}
    \pgfmathsetmacro\indext{3*\indexo-1}
    \pgfmathsetmacro\indextt{\indext-2}
    \ifnum\x=35
    	\node[] (\x) [above=of 19] {$\cdots$};
    \else
    	\ifnum\x<35
			\node[squarednode] (\x) [above=of \indext] {};
        \else
        	\node[squarednode] (\x) [above=of \indextt] {};
        \fi
    \fi    
}   
\foreach \x in {29,...,38}{ 
	\pgfmathsetmacro\indexo{\x-28}
    \pgfmathsetmacro\indext{3*\indexo-1}
    \pgfmathsetmacro\indextt{\indext-2}
            	\pgfmathtruncatemacro\lc{\indext-1}
            \pgfmathtruncatemacro\rc{\indext+1}
            \pgfmathtruncatemacro\lct{\indextt-1}
            \pgfmathtruncatemacro\rct{\indextt+1}
    \ifnum\x=35

    \else
    	\ifnum\x<35
        	\draw[-,black] (\x.south) -- (\lc.north);
            \draw[-,black] (\x.south) -- (\rc.north);
        \else
        	\draw[-,black] (\x.south) -- (\lct.north);
            \draw[-,black] (\x.south) -- (\rct.north);
        \fi
    \fi    
} 

\foreach \x in {39,41,42}{ 
	\pgfmathsetmacro\indexo{\x-38}
    \pgfmathtruncatemacro\indext{3*\indexo+27}
    \pgfmathtruncatemacro\indextt{\indext-2}
            	\pgfmathtruncatemacro\lc{\indext-1}
            \pgfmathtruncatemacro\rc{\indext+1}
            \pgfmathtruncatemacro\lct{\indextt-1}
            \pgfmathtruncatemacro\rct{\indextt+1}
    \ifnum\x<41
    		\node[squarednode] (\x) [above=of \indext] {};
        	\draw[-,black] (\x.south) -- (\lc.north);
            \draw[-,black] (\x.south) -- (\rc.north);
            \draw[-,black] (\x.south) -- (\indext.north);
    \else
    	\ifnum\x=41
        	\node[] (\x) [above=of 35] {$\cdots$};
        \else
        	\node[squarednode] (\x) [above=of \indextt] {};
            \draw[-,black] (\x.south) -- (\lct.north);
            \draw[-,black] (\x.south) -- (\rct.north);
            \draw[-,black] (\x.south) -- (\indextt.north);
        \fi
    \fi    
} 

\node[squarednode] (43) [above=of 32] {};
\draw[-,black] (43.south) -- (31.north);
\draw[-,black] (43.south) -- (32.north);
\draw[-,black] (43.south) -- (33.north);

\draw[-,black,dashed] (41.south) -- (36.north);

\node[squarednode] (45) [above=of 34] {};
\draw[-,black] (45.south) -- (33.north);
\draw[-,black] (45.south) -- (34.north);
\draw[-,black,dashed] (45.south) -- (35.north);

\draw[decoration={brace,mirror,raise=20pt},decorate] (1) -- node[below=21pt, font=\huge] {51 weeks} (28);
\end{tikzpicture}
}
\caption{CDEFs 1-51, 2-25}
\end{subfigure}
\hfill

\begin{subfigure}[h]{\textwidth}
\resizebox{\columnwidth}{!}{
\begin{tikzpicture}[
roundnode/.style={circle, draw=green!60, fill=green!5, very thick, minimum size=1mm},
squarednode/.style={circle, draw=blue!60, fill=red!5, very thick, minimum size=1mm},
]
\node[squarednode]      (1)                              {};
\node[squarednode]      (3)        [right=of 1]        {};
\coordinate (2) at ($(1)!0.5!(3)$);
\draw (2) node[] {$\cdots$};

\node[squarednode]      (4)        [right=of 3]                      {};
\node[squarednode]      (6)       [right=of 4]   {};
\coordinate (5) at ($(4)!0.5!(6)$);
\draw (5) node[] {$\cdots$};

\node[squarednode]      (7)        [right=of 6]        {};
\node[squarednode]      (9)        [right=of 7]        {};
\coordinate (8) at ($(7)!0.5!(9)$);
\draw (8) node[] {$\cdots$};

\node[squarednode]      (10)	[right=of 9]           {};
\node[squarednode]      (12)        [right=of 10]        {};
\coordinate (11) at ($(10)!0.5!(12)$);
\draw (11) node[] {$\cdots$};

\node[squarednode]      (13)        [right=of 12]                      {};
\node[squarednode]      (15)       [right=of 13]   {};
\coordinate (14) at ($(13)!0.5!(15)$);
\draw (14) node[] {$\cdots$};

\node[squarednode]      (16)        [right=of 15]        {};
\node[squarednode]      (18)        [right=of 16]        {};
\coordinate (17) at ($(16)!0.5!(18)$);
\draw (17) node[] {$\cdots$};

\node[]      (19)        [right=of 18]        {$\cdots$};

\node[squarednode]      (20)       [right=of 19]   {};
\node[squarednode]      (22)        [right=of 20]        {};
\coordinate (21) at ($(20)!0.5!(22)$);
\draw (21) node[] {$\cdots$};

\node[squarednode]      (23)       [right=of 22]   {};
\node[squarednode]      (25)        [right=of 23]        {};
\coordinate (24) at ($(23)!0.5!(25)$);
\draw (24) node[] {$\cdots$};

\node[squarednode]      (26)       [right=of 25]   {};
\node[squarednode]      (28)        [right=of 26]        {};
\coordinate (27) at ($(26)!0.5!(28)$);
\draw (27) node[] {$\cdots$};

\foreach \x in {29,...,38}{ 
	\pgfmathsetmacro\indexo{\x-28}
    \pgfmathsetmacro\indext{3*\indexo-1}
    \pgfmathsetmacro\indextt{\indext-2}
    \ifnum\x=35
    	\node[] (\x) [above=of 19] {$\cdots$};
    \else
    	\ifnum\x<35
			\node[squarednode] (\x) [above=of \indext] {};
        \else
        	\node[squarednode] (\x) [above=of \indextt] {};
        \fi
    \fi    
}   
\foreach \x in {29,...,38}{ 
	\pgfmathsetmacro\indexo{\x-28}
    \pgfmathsetmacro\indext{3*\indexo-1}
    \pgfmathsetmacro\indextt{\indext-2}
            	\pgfmathtruncatemacro\lc{\indext-1}
            \pgfmathtruncatemacro\rc{\indext+1}
            \pgfmathtruncatemacro\lct{\indextt-1}
            \pgfmathtruncatemacro\rct{\indextt+1}
    \ifnum\x=35

    \else
    	\ifnum\x<35
        	\draw[-,black] (\x.south) -- (\lc.north);
            \draw[-,black] (\x.south) -- (\rc.north);
        \else
        	\draw[-,black] (\x.south) -- (\lct.north);
            \draw[-,black] (\x.south) -- (\rct.north);
        \fi
    \fi    
} 

\foreach \x in {39,...,42}{ 
	\pgfmathsetmacro\indexo{\x-38}
    \pgfmathtruncatemacro\indext{3*\indexo+27}
    \pgfmathtruncatemacro\indextt{\indext-2}
            	\pgfmathtruncatemacro\lc{\indext-1}
            \pgfmathtruncatemacro\rc{\indext+1}
            \pgfmathtruncatemacro\lct{\indextt-1}
            \pgfmathtruncatemacro\rct{\indextt+1}
    \ifnum\x<41
    		\node[squarednode] (\x) [above=of \indext] {};
        	\draw[-,black] (\x.south) -- (\lc.north);
            \draw[-,black] (\x.south) -- (\rc.north);
            \draw[-,black] (\x.south) -- (\indext.north);
    \else
    	\ifnum\x=41
        	\node[] (\x) [above=of 35] {$\cdots$};
        \else
        	\node[squarednode] (\x) [above=of \indextt] {};
            \draw[-,black] (\x.south) -- (\lct.north);
            \draw[-,black] (\x.south) -- (\rct.north);
            \draw[-,black] (\x.south) -- (\indextt.north);
        \fi
    \fi    
} 

\node[squarednode] (43) [above=of 32] {};
\draw[-,black] (43.south) -- (31.north);
\draw[-,black] (43.south) -- (32.north);
\draw[-,black] (43.south) -- (33.north);

\node[squarednode] (44) [above=of 36] {};
\draw[-,black,dashed] (44.south) -- (35.north);
\draw[-,black] (44.south) -- (36.north);
\draw[-,black] (44.south) -- (37.north);

\node[squarednode] (45) [above=of 34] {};
\draw[-,black] (45.south) -- (33.north);
\draw[-,black] (45.south) -- (34.north);
\draw[-,black,dashed] (45.south) -- (35.north);

\node[squarednode] (46) [above=of 31] {};
\draw[-,black] (46.south) -- (30.north);
\draw[-,black] (46.south) -- (31.north);
\draw[-,black] (46.south) -- (32.north);

\draw[decoration={brace,mirror,raise=20pt},decorate] (1) -- node[below=21pt, font=\huge] {51 weeks} (28);
\end{tikzpicture}
}
\caption{CDEFs 1-51, 2-49}
\end{subfigure}
\caption{Model Structures}
\label{fig:cdefsstructurelocation}
\end{figure}

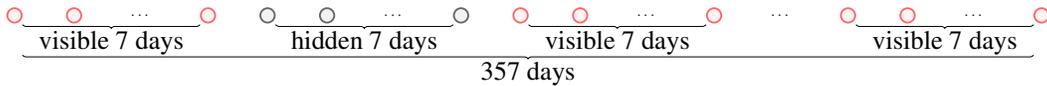
\begin{figure}[H]
\resizebox{\columnwidth}{!}{
\begin{tikzpicture}[
vnode/.style={circle, draw=red!60, fill=red!5, very thick, minimum size=1mm},
hnode/.style={circle, draw=black!60, fill=black!5, very thick, minimum size=1mm},
]
\foreach \x in {1,...,17}{
\pgfmathtruncatemacro\pre{\x-1}
	\ifnum\x=1
    	\node[vnode] (\x) {};
    \else
		\ifnum\x=3\relax
        	\node[] (\x) [right=of \pre] {$\cdots$};
        \else
        	\ifnum\x<5\relax
				\node[vnode] (\x) [right=of \pre] {};
        	\else
        		\ifnum\x<9
            		\ifnum\x=7
                		\node[] (\x) [right=of \pre] {$\cdots$};
                	\else
            			\node[hnode] (\x) [right=of \pre] {};
                 	\fi
            	\else
            	    \ifnum\x<13
                    	\ifnum\x=11
                    		\node[] (\x) [right=of \pre] {$\cdots$};
                        \else
                    		\node[vnode] (\x) [right=of \pre] {};
                    	\fi
                    \else
                    	\ifnum\x=13
                        	\node[] (\x) [right=of \pre] {$\cdots$};
                        \else
                        	\ifnum\x=16
                            	\node[] (\x) [right=of \pre] {$\cdots$};
                            \else
                            	\node[vnode] (\x) [right=of \pre] {};
                            \fi
                        \fi
                    \fi
                \fi
             \fi
    	\fi
    \fi
}

\draw[decoration={brace,mirror,raise=6pt},decorate]
  (1) -- node[below=7pt, font=\LARGE] {visible 7 days} (4);
\draw[decoration={brace,mirror,raise=6pt},decorate]
(5) -- node[below=7pt,font=\LARGE] {hidden 7 days} (8);
\draw[decoration={brace,mirror,raise=6pt},decorate]
(9) -- node[below=7pt,font=\LARGE] {visible 7 days} (12);
\draw[decoration={brace,mirror,raise=6pt},decorate]
(14) -- node[below=7pt,font=\LARGE] {visible 7 days} (17);
  \draw[decoration={brace,mirror,raise=26pt},decorate]
  (1) -- node[below=27pt,font=\LARGE] {357 days} (17);
\end{tikzpicture}
}
\caption{The representation of the data. Each node represents the 77 numbers of thefts at 77 locations for 1 day.}
\label{fig:datahidelocations}
\end{figure}

\noindent We ran the experiments 14 times. For each time, we chose a different year as the testing data. The other 13 years were the training data. We hid the data, as in Figure \ref{fig:datahidelocations}, for every other week in the testing year, \textit{i.e.}, we used the data in the odd number of weeks to estimate the number of thefts in the even number of weeks.

\begin{figure}[H]
\centering
\includegraphics[width=\textwidth]{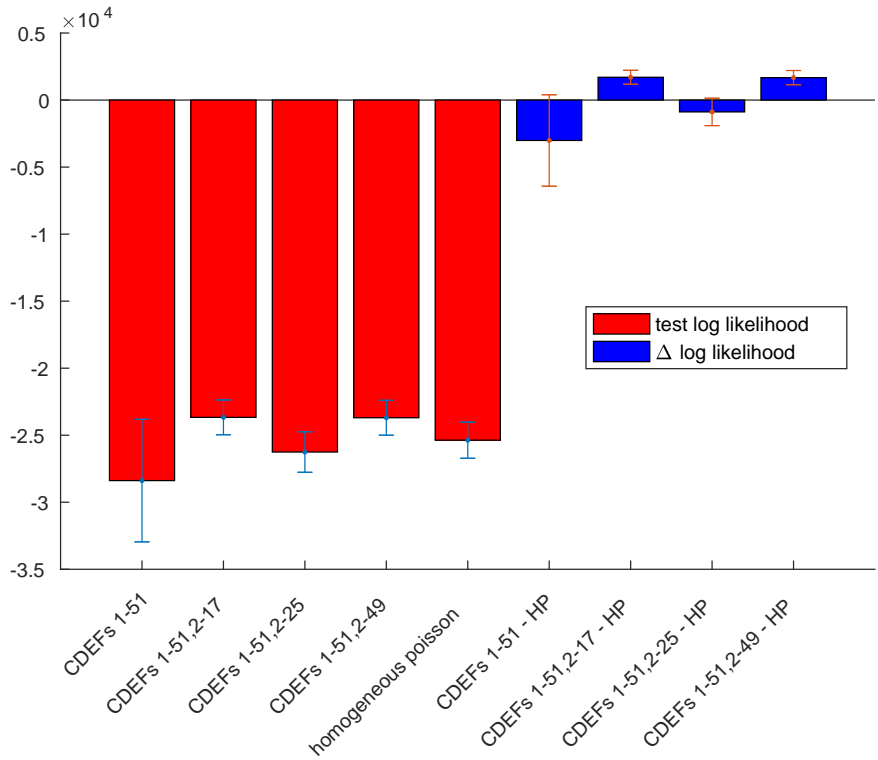}
\caption{The left 5 bars are the errorbars of test log likelihood for the above 5 models. The right 4 bars are the errorbars for the difference between the test log likelihood of CDEFs and HP at each run of the experiments.}
\label{fig:ex1}
\end{figure}

\noindent As shown in Figure \ref{fig:ex1}, CDEFs with 2 layers have larger test log likelihoods than 1-layer CDEFs. CDEFs 1-51, 2-17 and CDEFs 1-51, 2-49 perform better than homogeneous Poisson model. 
\subsection{The Benefits of The Overlap}
The overlap of the filter can give some clues for all the hidden nodes connected to it.  In this way, we could use the hidden nodes to capture the time dependence with less data.

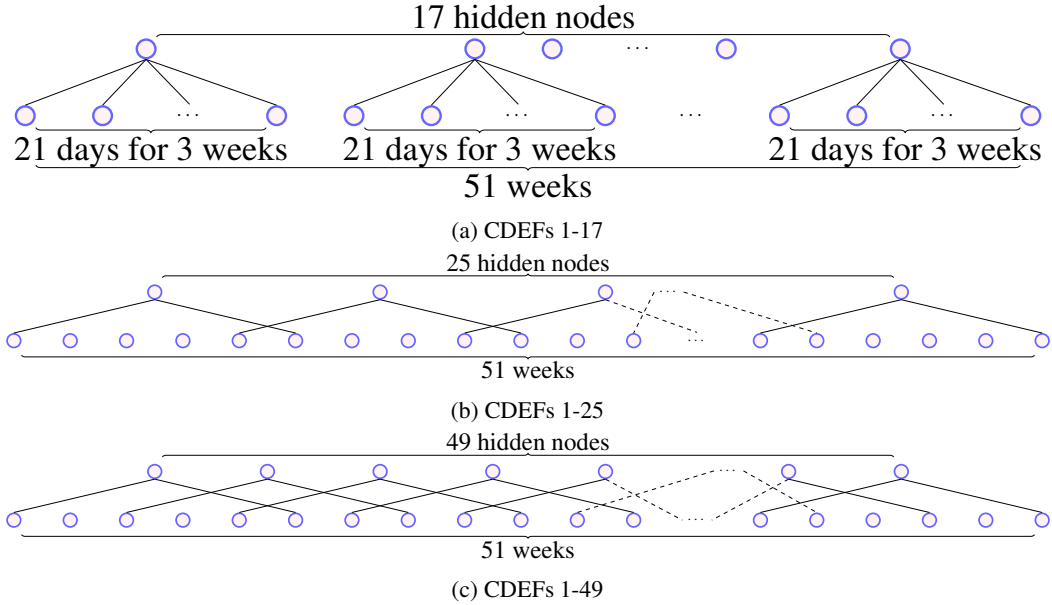
\begin{figure}[H]
\centering
\begin{subfigure}[h]{\textwidth}
\resizebox{\columnwidth}{!}{
\begin{tikzpicture}[
roundnode/.style={circle, draw=green!60, fill=green!5, very thick, minimum size=1mm},
squarednode/.style={circle, draw=blue!60, fill=red!5, very thick, minimum size=1mm},
]
\node[squarednode]      (1)                              {};
\node[squarednode]      (2)        [right=of 1]        {};
\node[]                 (3)        [right=of 2]        {$\cdots$};
\coordinate (Middle) at ($(2)!0.5!(3)$);
\node[squarednode]      (10)       [above=of Middle]   {};
\node[squarednode]      (4)        [right=of 3]        {};

\node[squarednode]      (16)        [right=of 4]                      {};
\node[squarednode]      (17)        [right=of 16]        {};
\node[]                 (18)        [right=of 17]        {$\cdots$};
\coordinate (Middle3) at ($(17)!0.5!(18)$);
\node[squarednode]      (19)       [right=of 18]   {};

\node[]                 (5)        [right=of 19]        {$\cdots$};
\node[squarednode]      (6)        [right=of 5]        {};
\node[squarednode]      (7)        [right=of 6]        {};
\node[]                 (8)        [right=of 7]        {$\cdots$};
\coordinate (Middle2) at ($(7)!0.5!(8)$);
\node[squarednode]      (11)       [above=of Middle2]   {};
\node[squarednode]      (9)        [right=of 8]        {};

\node[squarednode]      (12)        [above=of Middle3]        {};
\node[squarednode]      (13)        [right=of 12]        {};
\node[]      (14)        [right=of 13]        {$\cdots$};
\node[squarednode]      (15)        [right=of 14]        {};

\draw[-,black] (10.south) -- (1.north);
\draw[-,black] (10.south) -- (2.north);
\draw[-,black] (10.south) -- (3.north);
\draw[-, black] (10.south) -- (4.north);

\draw[-,black] (11.south) -- (6.north);
\draw[-,black] (11.south) -- (7.north);
\draw[-,black] (11.south) -- (8.north);
\draw[-, black] (11.south) -- (9.north);

\draw[-,black] (12.south) -- (16.north);
\draw[-,black] (12.south) -- (17.north);
\draw[-,black] (12.south) -- (18.north);
\draw[-, black] (12.south) -- (19.north);

\draw[decoration={brace,mirror,raise=6pt},decorate]
  (1) -- node[below=7pt,font=\LARGE] {21 days for 3 weeks} (4);
\draw[decoration={brace,mirror,raise=6pt},decorate]
(6) -- node[below=7pt,font=\LARGE] {21 days for 3 weeks} (9);
\draw[decoration={brace,mirror,raise=6pt},decorate]
(16) -- node[below=7pt,font=\LARGE] {21 days for 3 weeks} (19);
\draw[decoration={brace,raise=6pt},decorate]
(10) -- node[above=7pt,font=\LARGE] {17 hidden nodes} (11);
  \draw[decoration={brace,mirror,raise=25pt},decorate]
  (1) -- node[below=26pt,font=\LARGE] {51 weeks} (9);
\end{tikzpicture}
}
\caption{CDEFs 1-17}
\end{subfigure}
\hfill
\begin{subfigure}[h]{\textwidth}
\resizebox{\columnwidth}{!}{
\begin{tikzpicture}[
roundnode/.style={circle, draw=green!60, fill=green!5, very thick, minimum size=1mm},
squarednode/.style={circle, draw=blue!60, fill=red!5, very thick, minimum size=1mm},
]
\node[squarednode] (1) {};
\foreach \x in {2,...,19}{
	\pgfmathtruncatemacro\pre{\x-1}
    \ifnum\x=13
    	\node[] (\x) [right=of \pre] {$\cdots$};
    \else
    	\node[squarednode] (\x) [right=of \pre] {};
    \fi
}

\coordinate (middle1) at ($(3)!0.5!(4)$);
\node[squarednode] (20) [above=of middle1] {};
\draw[-,black] (20.south) -- (1.north);
\draw[-,black] (20.south) -- (6.north);

\coordinate (middle2) at ($(7)!0.5!(8)$);
\node[squarednode] (21) [above=of middle2] {};
\draw[-,black] (21.south) -- (5.north);
\draw[-,black] (21.south) -- (10.north);

\coordinate (middle3) at ($(11)!0.5!(12)$);
\node[squarednode] (22) [above=of middle3] {};
\draw[-,black] (22.south) -- (9.north);
\draw[-,black,dashed] (22.south) -- (13.north);

\node[] (23) [right=of 22] {$\cdots$};
\draw[-,black,dashed] (23.west) -- (12.north);
\draw[-,black,dashed] (23.east) -- (15.north);

\coordinate (middle4) at ($(16)!0.5!(17)$);
\node[squarednode] (24) [above=of middle4] {};
\draw[-,black] (24.south) -- (14.north);
\draw[-,black] (24.south) -- (19.north);

\draw[decoration={brace,raise=10pt},decorate]
(20) -- node[above=11pt,font=\LARGE] {25 hidden nodes} (24);
\draw[decoration={brace,mirror,raise=10pt},decorate]
  (1) -- node[below=11pt,font=\LARGE] {51 weeks} (19);
\end{tikzpicture}
}
\caption{CDEFs 1-25}
\end{subfigure}
\hfill
\begin{subfigure}[h]{\textwidth}
\resizebox{\columnwidth}{!}{
\begin{tikzpicture}[
roundnode/.style={circle, draw=green!60, fill=green!5, very thick, minimum size=1mm},
squarednode/.style={circle, draw=blue!60, fill=red!5, very thick, minimum size=1mm},
]
\node[squarednode] (1) {};
\foreach \x in {2,...,19}{
	\pgfmathtruncatemacro\pre{\x-1}
    \ifnum\x=13
    	\node[] (\x) [right=of \pre] {$\cdots$};
    \else
    	\node[squarednode] (\x) [right=of \pre] {};
    \fi
}
\coordinate (middle1) at ($(3)!0.5!(4)$);
\node[squarednode] (20) [above=of middle1] {};
\draw[-,black] (20.south) -- (1.north);
\draw[-,black] (20.south) -- (6.north);

\coordinate (middle2) at ($(5)!0.5!(6)$);
\node[squarednode] (21) [above=of middle2] {};
\draw[-,black] (21.south) -- (3.north);
\draw[-,black] (21.south) -- (8.north);

\coordinate (middle3) at ($(7)!0.5!(8)$);
\node[squarednode] (22) [above=of middle3] {};
\draw[-,black] (22.south) -- (5.north);
\draw[-,black] (22.south) -- (10.north);

\coordinate (middle4) at ($(9)!0.5!(10)$);
\node[squarednode] (23) [above=of middle4] {};
\draw[-,black] (23.south) -- (7.north);
\draw[-,black] (23.south) -- (12.north);

\coordinate (middle5) at ($(11)!0.5!(12)$);
\node[squarednode] (24) [above=of middle5] {};
\draw[-,black] (24.south) -- (9.north);
\draw[-,black,dashed] (24.south) -- (13.west);

\coordinate (middle6) at ($(13)!0.5!(14)$);
\node[] (25) [above=of middle6] {$\cdots$};
\draw[-,black,dashed] (25.west) -- (11.north);

\draw[-,black,dashed] (25.east) -- (15.north);

\coordinate (middle7) at ($(14)!0.5!(15)$);
\node[squarednode] (26) [above=of middle7] {};
\draw[-,black] (26.south) -- (17.north);
\draw[-,black,dashed] (26.south) -- (13.east);

\coordinate (middle8) at ($(16)!0.5!(17)$);
\node[squarednode] (27) [above=of middle8] {};
\draw[-,black] (27.south) -- (14.north);
\draw[-,black] (27.south) -- (19.north);

\draw[decoration={brace,mirror,raise=10pt},decorate]
  (1) -- node[below=11pt,font=\LARGE] {51 weeks} (19);
  \draw[decoration={brace,raise=10pt},decorate]
(20) -- node[above=11pt,font=\LARGE] {49 hidden nodes} (27);
\end{tikzpicture}
}
\caption{CDEFs 1-49}
\end{subfigure}
\caption{Model Structures}
\label{fig:cdefsex2}
\end{figure}

We constructed 3 CDEFs models as in Figure \ref{fig:cdefsex2}:
\begin{itemize}
\item CDEFs 1-17: 1 layer CDEFs with 17 hidden nodes. Each hidden node is connected with 3 weeks of data. The filter size is 1617 and the stride is 1617 ($1617 = 3 \text{ weeks } \times 7 \text{ days } \times 77 \text{ locations }$). There is no overlap.
\item CDEFs 1-25: 1 layer CDEFs with 25 hidden nodes. Each hidden node is connected with 3 weeks of data. The filter size is 1617 and the stride is 1078 ( the number of data points for 2 weeks ). 
\item CDEFs 1-49: 1 layer CDEFs with 49 hidden nodes. Each hidden node is connected with 3 weeks of data. The filter size is 1617 and the stride is 539 ( the number of data points for 1 week ). 
\end{itemize}

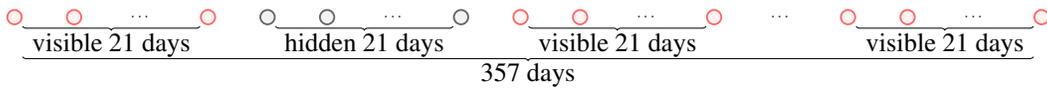
\begin{figure}[H]
\resizebox{\columnwidth}{!}{
\begin{tikzpicture}[
vnode/.style={circle, draw=red!60, fill=red!5, very thick, minimum size=1mm},
hnode/.style={circle, draw=black!60, fill=black!5, very thick, minimum size=1mm},
]
\foreach \x in {1,...,17}{
\pgfmathtruncatemacro\pre{\x-1}
	\ifnum\x=1
    	\node[vnode] (\x) {};
    \else
		\ifnum\x=3\relax
        	\node[] (\x) [right=of \pre] {$\cdots$};
        \else
        	\ifnum\x<5\relax
				\node[vnode] (\x) [right=of \pre] {};
        	\else
        		\ifnum\x<9
            		\ifnum\x=7
                		\node[] (\x) [right=of \pre] {$\cdots$};
                	\else
            			\node[hnode] (\x) [right=of \pre] {};
                 	\fi
            	\else
            	    \ifnum\x<13
                    	\ifnum\x=11
                    		\node[] (\x) [right=of \pre] {$\cdots$};
                        \else
                    		\node[vnode] (\x) [right=of \pre] {};
                    	\fi
                    \else
                    	\ifnum\x=13
                        	\node[] (\x) [right=of \pre] {$\cdots$};
                        \else
                        	\ifnum\x=16
                            	\node[] (\x) [right=of \pre] {$\cdots$};
                            \else
                            	\node[vnode] (\x) [right=of \pre] {};
                            \fi
                        \fi
                    \fi
                \fi
             \fi
    	\fi
    \fi
}

\draw[decoration={brace,mirror,raise=6pt},decorate]
  (1) -- node[below=7pt, font=\LARGE] {visible 21 days} (4);
\draw[decoration={brace,mirror,raise=6pt},decorate]
(5) -- node[below=7pt,font=\LARGE] {hidden 21 days} (8);
\draw[decoration={brace,mirror,raise=6pt},decorate]
(9) -- node[below=7pt,font=\LARGE] {visible 21 days} (12);
\draw[decoration={brace,mirror,raise=6pt},decorate]
(14) -- node[below=7pt,font=\LARGE] {visible 21 days} (17);
  \draw[decoration={brace,mirror,raise=26pt},decorate]
  (1) -- node[below=27pt,font=\LARGE] {357 days} (17);
\end{tikzpicture}
}
\caption{The representation of the data.}
\label{fig:datahidelocationsex2}
\end{figure}

\begin{figure}[H]
\includegraphics[width=\textwidth]{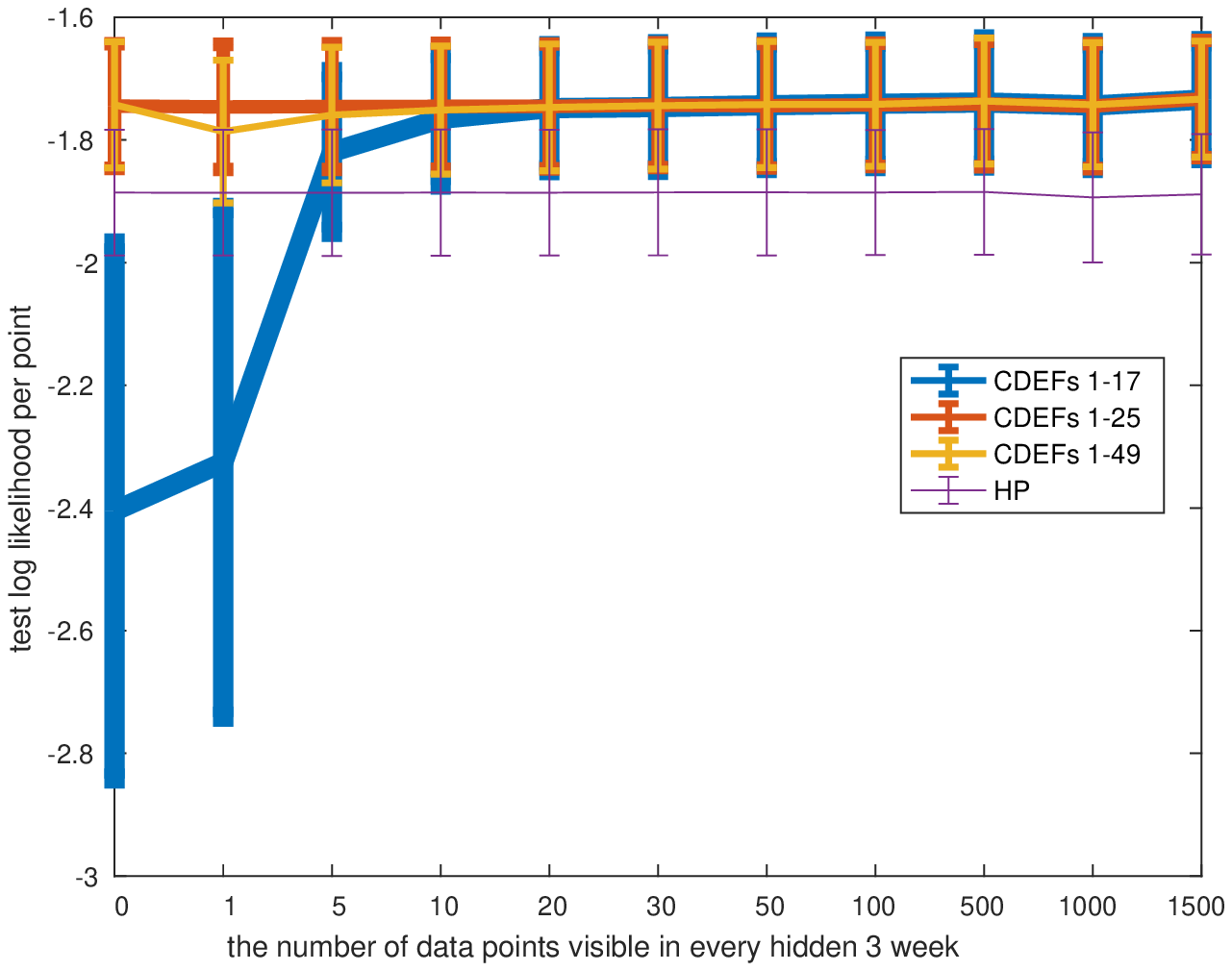}
\caption{Results}
\label{fig:ex2}
\end{figure}

First, we hid every other 3 weeks of data as in Figure \ref{fig:datahidelocationsex2}. Then, we kept increasing the number of data points visible in every hidden 3 weeks. The visible points in the hidden 3 weeks were chosen randomly. We also ran the experiments 14 times, the same as the previous one.
As shown in Figure \ref{fig:ex2}, when there is no or very small number of data points visible in the hidden 3 weeks, the overlapping CDEFs (CDEFs 1-25, CDEFs 1-49) behave better than non-overlapping CDEFs (CDEFs 1-17) and homogeneous Poisson process. It shows that the CDEFs with overlap can utilize the data better since the overlap can help update the parameters of the hidden nodes connected to it. As the number of visible data points increases, the performance of CDEFs 1-17 becomes better at first and then has almost the same behavior as the other CDEFs models, and there is no significant improvement for Homogeneous Poisson process or the overlapping CDEFs.  The reason is that the number of thefts is very similar for each location at different days. Even the number of visible points increases, the estimation of the rates of Poisson distribution remains almost the same.
\section{Conclusion}
In this paper, we developed convolutional deep exponential families as an efficient way to capture time correlation. We have also designed some experiments to show how CDEFs with deep structure or overlap behave well with small amount of data.



\bibliography{ref}










\end{document}